\documentclass[letterpaper, 10 pt, conference]{ieeeconf}
\IEEEoverridecommandlockouts % This command is only needed if you want to use the \thanks command
\usepackage{cite}
\usepackage{amsmath,amssymb,amsfonts}
\usepackage{algorithm} 
\usepackage{algpseudocode} 
\usepackage{graphicx}
\usepackage{textcomp}
\usepackage{xcolor}
\usepackage{caption}
\usepackage{subcaption}
\usepackage{float}

\usepackage{booktabs}
\usepackage{multirow}
\title{\LARGE \bf Concept-based Anomaly Detection in Retail Stores for Automatic Correction using Mobile Robots}

\author{Aditya Kapoor$^{1}$, Vartika Sengar$^{1}$, Nijil George$^{1}$, Vighnesh Vatsal$^{1}$  \\Jayavardhana Gubbi$^{1}$, Balamuralidhar P$^{1}$ and Arpan Pal$^{1}$
\thanks{$^{1}$The authors are with TCS Research, Tata Consultancy Services Ltd., Bengaluru, Karnataka - 560066, India. e-mails: \{ \tt\small aditya.kapoor1, vartika.sengar, george.nijil, vighnesh.vatsal, j.gubbi,  balamurali.p, arpan.pal\} @tcs.com}
}

\begin{document}

\maketitle

\begin{abstract}
Tracking of inventory and rearrangement of misplaced items are some of the most labor-intensive tasks in a retail environment. While there have been attempts at using vision-based techniques for these tasks, they mostly use planogram compliance for detection of any anomalies, a technique that has been found lacking in robustness and scalability. Moreover, existing systems rely on human intervention to perform corrective actions after detection. In this paper, we present Co-AD, a Concept-based Anomaly Detection approach using a Vision Transformer (ViT) that is able to flag misplaced objects without using a prior knowledge base such as a planogram. It uses an auto-encoder architecture followed by outlier detection in the latent space. Co-AD has a peak success rate of 89.90\% on anomaly detection image sets of retail objects drawn from the RP2K dataset, compared to 80.81\% on the best-performing baseline of a standard ViT auto-encoder. To demonstrate its utility, we describe a robotic mobile manipulation pipeline to autonomously correct the anomalies flagged by Co-AD. This work is ultimately aimed towards developing autonomous mobile robot solutions that reduce the need for human intervention in retail store management.  
\end{abstract}

% \begin{IEEEkeywords}
% Perception, Anomaly detection, Concept mining, Mobile manipulator robots
% \end{IEEEkeywords}

\section{Introduction}
\label{sec:intro}

In recent years, there have been significant technological innovations in retail store management, focusing on supply chains, logistics, and inventory tracking.
With demographic shifts in industrialized economies, there is a push towards automating repetitive, labor-intensive tasks in domains such as retail stores and supermarkets.
Advancements in robotics have enabled the possibility of deploying mobile manipulator robots for performing these tasks efficiently and autonomously.

One of the key activities in the retail domain is planogram compliance. A planogram is a schematic diagram predefined by store operators, laying out the shelf-wise location of each product, aimed at maximizing visibility and sales potential.
Compliance implies checking the current layout and ensuring that it adheres to the planogram by correcting any deviations from it.
Anomalies in the context of this paper refer to such deviations from the planogram.
Most planogram compliance solutions depend on recognition of individual product categories followed by comparison with the reference planogram to detect anomalies.
These solutions call for at least one reference image per product and a broad layout of how things should be placed~\cite{20}. 
Such solutions are reliant on manual updates as new product varieties are constantly introduced and shop layouts are updated often. 
Also, there may be large variations in the kinds of planograms that are available at a given retail store. 
For instance, a planogram may simply contain representative shapes or gray scale images corresponding to each product.
This negatively affects the performance of systems that rely solely on planogram matching to check for anomalies.
%Therefore, these solutions which use recognition with planogram matching may not perform as expected. 
To address this challenge, we propose Concept-based Anomaly Detection (Co-AD) which is a method for detecting out-of-context products in a retail environment. 
We also demonstrate how Co-AD can be a key component in developing mobile manipulation solutions that autonomously correct any flagged anomalies. 
\begin{figure}[t]
\centering
\includegraphics[width=0.52\columnwidth]{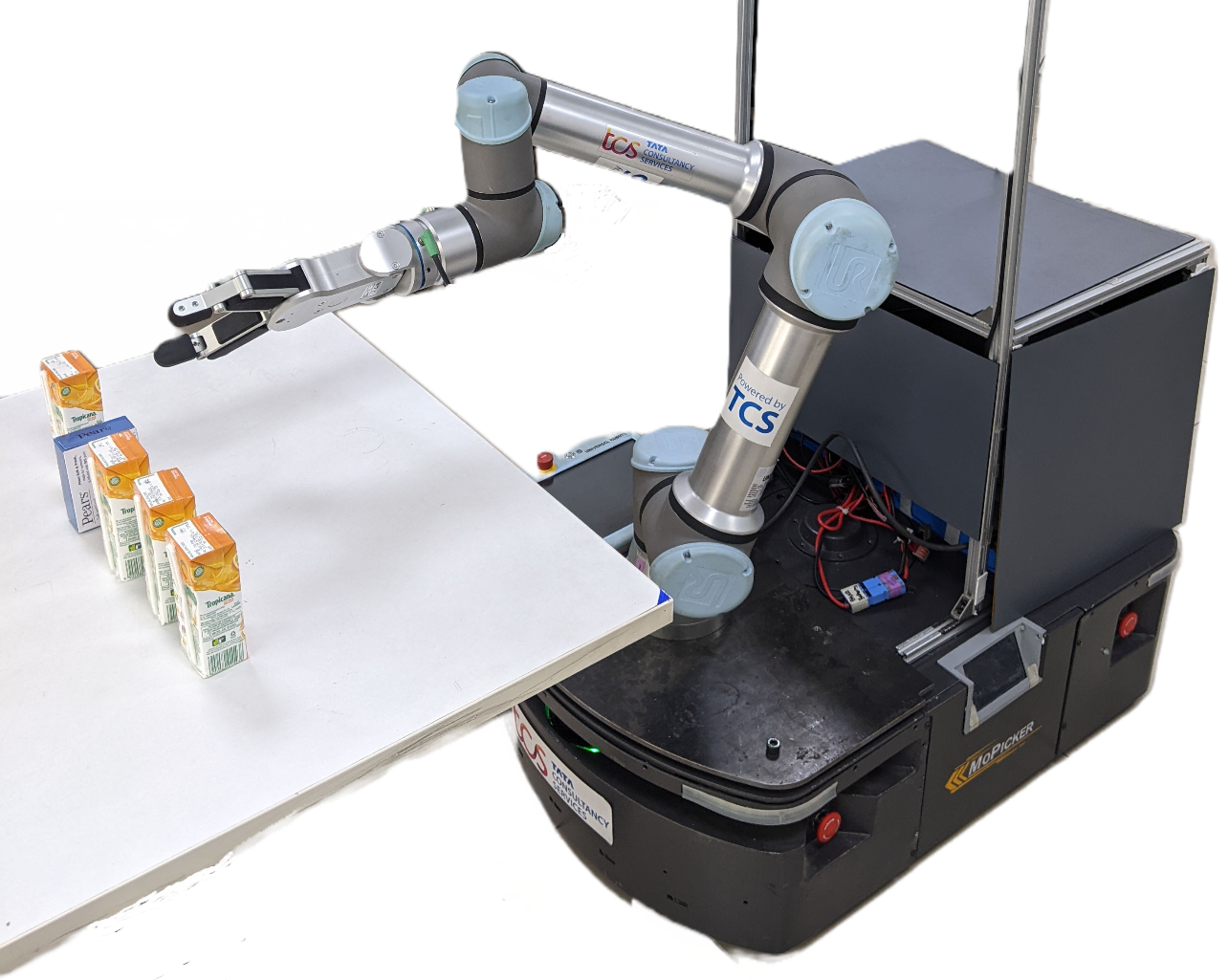}
\caption{Concept-based Anomaly Detection (Co-AD) can act as a key element in autonomous anomaly correction pipelines using mobile manipulator robots.}
\label{fig:intro_fig}
\end{figure}
\subsection{Related Work}
%Planogram compliance Papers
Current state-of-the-art solutions for autonomous retail store shelf management through computer vision typically solve the problem of detecting anomalies in a two-stage manner. The anomalies are flagged by checking whether or not the observed image is in agreement with the pre-defined planogram~\cite{17,18,19,20,21,22}. In the first stage, each product placed on the shelf is localized and classified. The recognition is done using either traditional hand-crafted feature descriptors such as key-points, gradients, patterns, colors, or feature embeddings acquired from deep learning (DL) models~\cite{3}. In ~\cite{17}, the products are detected based on different methods such as sliding window-based HOG (Histogram of Oriented Gradients) and BOVW (Bag of Visual Words) features. Some solutions~\cite{18,19,20} use SURF, SIFT and Hough transform-based features for detection. The other solutions for planogram compliance use template matching based on morphological gradients~\cite{21} and recurrent pattern recognition~\cite{22}. In the second stage, matching is done between the observations of the first stage and the actual layout given by the planogram. The methods used for matching include sub-graph isomorphism~\cite{20} and spectral graph matching~\cite{22}.

However, the dependence on planograms limits the use of such systems in retail settings due to periodic layout updates and constant introduction of anomalies by human agents (store visitors). Also, there are challenges arising from accurate object detection and recognition in cluttered scenes, classification with extremely high number of object classes, and adaptability towards new classes~\cite{wei2020deep}.
These difficulties, along with class imbalances, large label spaces, and low inter-class variance limit the success of these methods. 

In terms of robotic solutions, there have been mobile robots deployed in large retail stores for inventory tracking~\cite{simbe_patent}, using cameras mounted on a mobile base~\cite{Simbe,Zippedi}.
These robots are designed to detect missing items on shelves, and alert store employees so that they may take corrective action.  
However, the reliance on human intervention along with the aforementioned limitations of planogram compliance-based approaches affect the scalability and profitability of these systems. 

A possible solution to these shortcomings involves using scalable perception methods such as Co-AD, together with mobile manipulators to correct anomalies as they are detected, minimizing human intervention.
In this paper, we focus mainly on perception---detecting the anomalies, while demonstrating a correction pipeline using existing techniques for task planning, motion planning and grasping.
Another line of recent work in this field focuses on reactive action planning for mobile manipulators given a target object and goal state~\cite{pezzato2020active}.
Our ongoing and future work is in this domain as well, using our custom mobile base~\cite{mopicker_patent} with a manipulator arm as a research platform (Fig.~\ref{fig:intro_fig}) for autonomous task and motion planning in retail scenarios. 
\subsection{Contributions and Overview}
The key contributions of this paper are:
\begin{itemize}
    \item A concept-based anomaly detection method (Co-AD) for out-of-context objects in a retail environment. This method does not rely on planogram compliance.
    \item A demonstration of how such an anomaly detection technique can drive a robotics pipeline involving mobile manipulation to correct the anomalies. 
\end{itemize}

In Sec.~\ref{sec:system_desc}, we define the problem statement and the overall robotics pipeline.
In Sec.~\ref{sec:anom_det}, we describe the Co-AD algorithm in detail.
In Sec.~\ref{sec:results}, we find that Co-AD has a peak success rate of 89.9 \% on a retail image dataset (RP2K~\cite{rp2k}), and 95.83 \% on simulated images (YCB~\cite{ycb-dataset}).
Sec.~\ref{sec:demo} includes a demonstration of the application of Co-AD on mobile manipulation pipelines in simulated and real-world settings, followed by conclusions and future outlook in Sec.~\ref{sec:conclusion}.

\section{System Description}
\label{sec:system_desc}

\begin{figure*}[ht]
\centering
\includegraphics[width=0.85\textwidth]{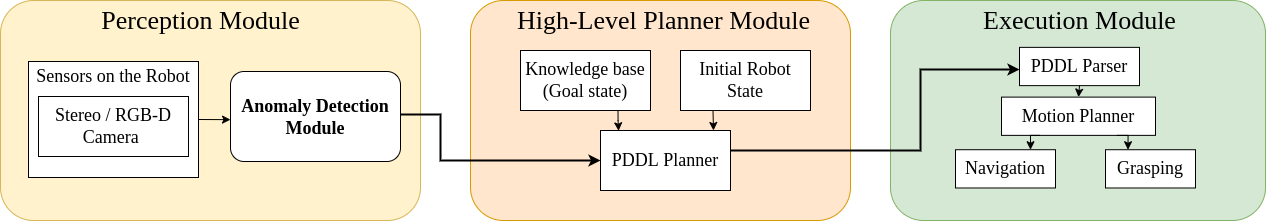}
\caption{The proposed solution pipeline for autonomous anomaly correction in a retail store.}
\label{fig:schematic}
\end{figure*}

\subsection{Anomalies in the Retail Setting}
Anomalies in retail settings are particularly different from the ones that have been tackled by current computer vision approaches. In the computer vision literature, anomalies can be classified as spatial or temporal, based on the type of input data~\cite{chalapathy2019deep}. In case of temporal anomalies, the goal is to detect the anomalous activity in a video. In spatial anomalies, the goal is to find an image which deviates from the distribution of a normal class. For example, for the CIFAR10~\cite{krizhevsky2009learning}, MNIST~\cite{lecun1998gradient} and ImageNet~\cite{krizhevsky2017imagenet} datasets, anomaly detection problems are posed as One-Class Novelty Detection where one class is considered as the normal class and the remaining classes are treated as anomalous~\cite{an2015variational, chalapathy2018anomaly, ruff2018deep}. Another example is defect detection~\cite{bergmann2019mvtec}, where the aim is to detect and localize spatial anomalies like cracks and structural changes in given images. 
% In this paper, our focus is on spatial anomalies, however, 
% but anomalies in retail do not fall under the umbrella of intra-object anomalies like crack-detection~\cite{bergmann2019mvtec} but rather lie in the inter-object umbrella. More so, 
% unlike 
The spatial anomalies discussed so far have a well defined distribution of normal class, however in our case the anomalies are contextual, making the problem of anomaly detection difficult. To ease some of these difficulties, we propose the following four types for categorizing anomalies in retail settings--

\subsubsection{Misplaced}
A \textit{misplaced item anomaly} occurs when an object of type \textit{P} has been incorrectly placed on a shelf row that has majority of type \textit{Q} objects. The intervention would be to move the object \textit{P} to either a region containing other type \textit{P} objects, or to move it to an inventory buffer location such as a backroom or storage area. 

\subsubsection{Out-of-stock}
An \textit{out-of-stock or missing-item anomaly} occurs when several objects of type \textit{P} are unavailable on a shelf row or the shelf row is empty. The intervention, in case of an empty shelf, would be to report that \textit{shelf is vacant} else restock the objects \textit{P} back to the shelf row from the inventory buffer location such as a backroom or storage area.

% \subsubsection{Missing item}
% A \textit{missing item anomaly} occurs few object of type \textit{P} above a certain threshold are unavailable on a shelf row that is meant for them. The intervention would be to restock the objects \textit{P} back to the shelf row from the inventory buffer location such as a backroom or storage area.

\subsubsection{Misalignment}
A \textit{misalignment anomaly} occurs when one or several objects of type \textit{P} are found in undesirable orientations. The intervention would be to re-align the objects \textit{P} back to their original orientation on the shelf row by inferring the original orientation from the already existing group of objects of type \textit{P}.

\subsubsection{Rearrangement}
A \textit{rearrangement anomaly} occurs when one or several objects of type \textit{P} and \textit{Q} are shuffled on a shelf that is meant for them. The intervention would be to re-arrange or regroup the objects \textit{P} and \textit{Q} back to their original positions on the shelf by inferring the original positions by gauging the existing object setup (based on count).

\subsection{Problem Formulation}
As discussed in Sec.~\ref{sec:intro}, any disruptions to a retail shelf arrangement that may require intervention from employees, are encapsulated under the umbrella term of ``anomalies".
In a real large-scale store, it would not be feasible to enumerate every instance of the possible anomaly types and generate relevant plans for the robot to correct them.
Based on feedback from retail store employees, particularly supermarkets, we restrict the scope of this work to one of the most commonly encountered anomalies--- misplaced items.

% Formally, a misplaced item anomaly occurs when an object of type \textit{A} has been incorrectly placed on a shelf row that is meant for objects of type \textit{B}. The intervention would be to move the object \textit{A} to either the shelf row or region containing other type \textit{A} objects, or to move it to an inventory buffer location such as a backroom or storage area.

The problem formulation for carrying out an intervention using a mobile manipulator therefore involves the following steps:

\begin{itemize}
\item Scan single rows in a shelf to detect anomalous object(s). Identification or categorization of the object is not necessary to detect a misplaced item. 
\item Identify the misplaced item. Determine its goal position in the store using prior knowledge (planogram or group locations or buffer zone in a store room).
\item Autonomously grasp the object, plan a path, navigate and place the object at its goal location.
\end{itemize}

\subsection{Solution Pipeline}
We propose a modular framework that would allow the robot to achieve the aforementioned steps (Fig.~\ref{fig:schematic}). We broadly categorize the anomaly detection and correction problem into the following three modules, allowing for incremental future scaling of each step of the solution:
\begin{itemize}
\item Anomaly Detection using \textbf{Perception} 
\item Planning to correct the Anomaly using a \textbf{High-Level Task Planner}
\item Executing the plan using \textbf{Motion Planning}
\end{itemize}

By doing so, it becomes easier to replace the algorithms in a module with another whereas an end-to-end approach does not offer such flexibility since it requires retraining or fine-tuning the the entire system. The anomaly detection module within the perception module and its inter-play with the other modules is the focus of this paper.\\

\subsubsection{Perception Module}
The \textit{Perception Module} perceives the environment around the robot with the help of sensors and uses this information to realise the anomalous object. In our current approach, we take a snapshot of the entire shelf and process it row by row. At first, we localize all the objects in the same shelf row (in simulation we directly use the ground truth whereas one can rely on prior works like \cite{shelf_row_detection_2014}). Then we localize each object in the shelf row using Yolo-v6 \cite{li2022yolov6} and pass the cropped images of them to a feature extraction module which produces the latent concept embeddings corresponding to each object. These embeddings will be used to identify the anomalous object. 

\subsubsection{High-Level Planner Module}
The \textit{High-Level Planner Module} takes inputs from the \textit{Perception Module} about the anomalous object and the existing \textit{Knowledge Base} (eg: its target location) and synthesizes a plan that enables the robot to correct the anomalies. In our current approach, we use PDDL~\cite{pddl1998} to plan and schedule the sequence of subroutines that would allow the robot to move and manipulate the anomalous object for correction.  

\subsubsection{Execution Module}
The \textit{Execution} module comprises of lower-level task primitives such as autonomous navigation and grasping that enable the robot to interact with the environment and correct any anomalies. In our current approach we use the ROS MoveIt! package~\cite{moveit} for motion planning which uses state-of-the-art inverse kinematics solvers, path planning algorithms, and collision detection. In case of grasping, we plan an inverse-kinematic path for the robotic arm's end-effector in joint space, placing it in front of the target object, followed by the execution of a two-fingered grasp.

All of these modules are integrated with one another using the ROS framework.

\section{Anomaly Detection Module}
\label{sec:anom_det}
The Co-AD approach consists of a number of components, which we will describe in order: product localization, disentangled concept embedding extraction and finally the full algorithm that arrives at anomaly detection and localization decisions. 

\begin{figure}[ht]

\begin{subfigure}[b]{0.49\columnwidth}
         \centering
         \includegraphics[width=0.99\textwidth]{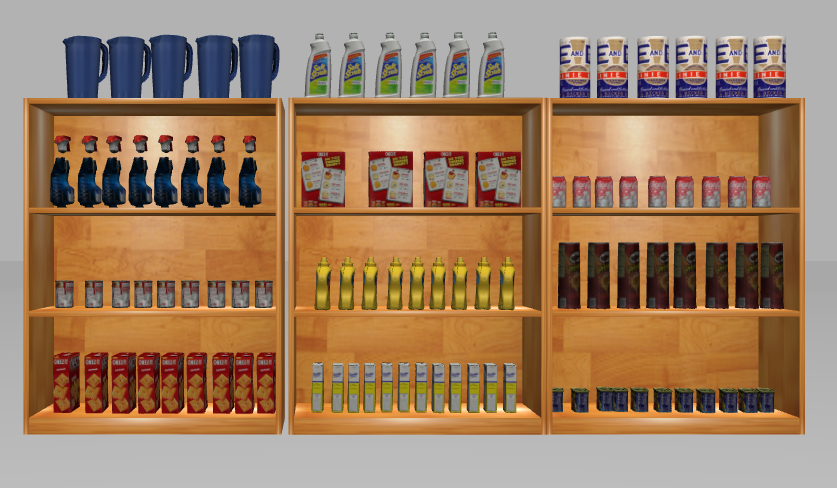}
         \caption{Simulated shelves}
         \label{fig:sub:gazebo_img}
\end{subfigure}
\begin{subfigure}[b]{0.49\columnwidth}
         \centering
         \includegraphics[width=0.9\textwidth]{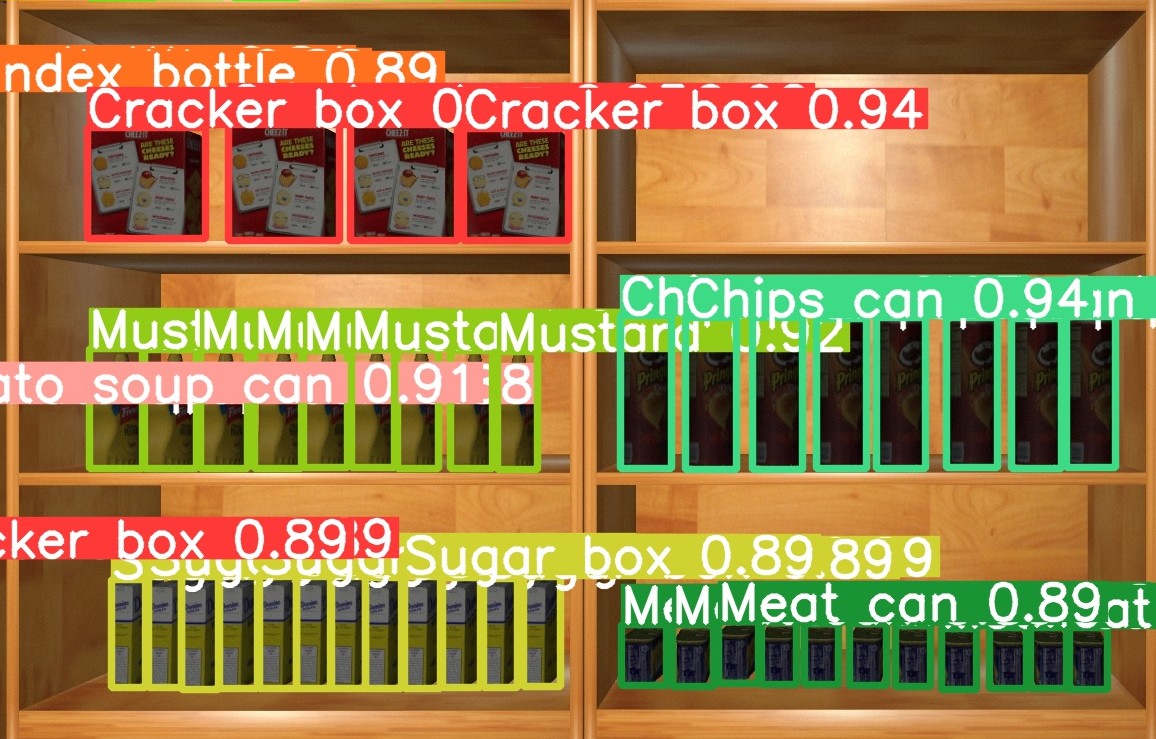}
         \caption{YOLO-v6 Results}
         \label{fig:sub:yolo_result}
\end{subfigure}
\caption{Simulation setup in Gazebo: (a) Simulated shelves, (b) The localization results of YOLO-v6.}
\label{fig:shelf_localization}
\end{figure}

\begin{figure*}[ht]
\centering
\includegraphics[width=0.9\textwidth]{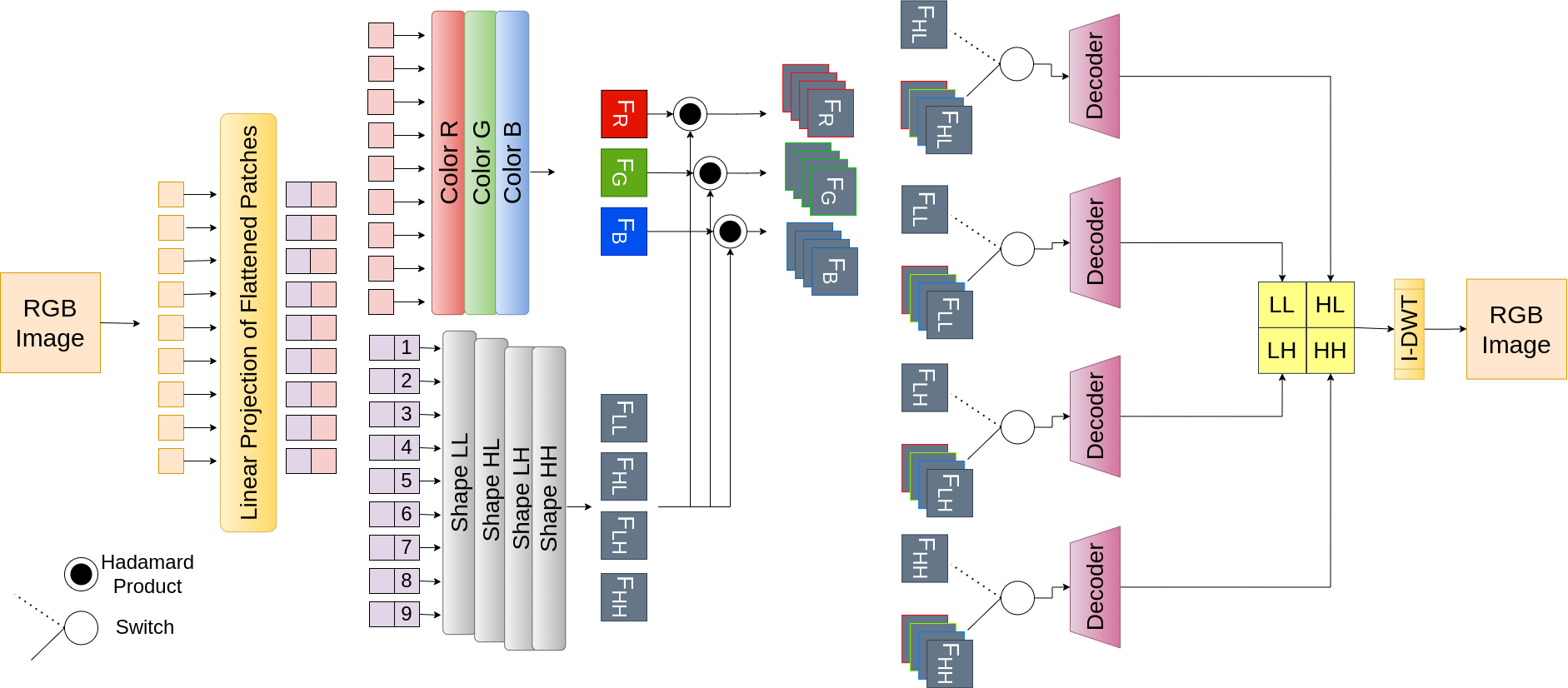}
\caption{The Concept Embedding Extraction Module Architecture.}
\label{fig:architecture}
\end{figure*}

\subsection{Product Localization Module}

The first module uses a single-stage deep learning detector to localize retail products. We use a YOLO-v6 \cite{li2022yolov6} detector to perform localization of different products in a given shelf image. The input to this module is pre-processed shelf images. 
% \textbf{How is the row in the shelf extracted}. 
The detector is trained on the SKU110k dataset \cite{goldman2019precise} for 400 epochs and batch size of 16 on a Tesla V100 GPU machine. 

Let us assume that $I_s$ represents the shelf image which contains several products (e.g. Fig.~\ref{fig:sub:gazebo_img}). The problem is to find the location of all the products in the given shelf image by rows. The localization module takes $I_s$ as input and gives $N_s$ bounding boxes, denoted by $b_{ij}$ where $j$ ranges from $1$ to $N_{sr}$ and $i$ ranges from $1$ to $N_{s}$. Here, $N_{sr}$ is the number of shelf rows and $N_{s}$ is the total number of objects. The localization results are shown in Fig.~\ref{fig:sub:yolo_result}. In simulation, we directly use the ground truth to uniquely identify the group of objects that are placed in different shelf rows after the localization step. Further, we crop the product images based on $b_{ij}$, which results in $I_c$ which is a set of cropped images. We process all the cropped images row-wise i.e., all cropped images belonging to a row $j$ are fed into a feature extraction module which outputs the concept latent embeddings for each object localized and is used for anomaly detection. This process is repeated for all the shelf rows. %In the next section, we'll explain details of concept embedding extraction module. 

\subsection{Concept Embedding Extraction Module}

Assume that a retail product dataset contains images of different products.
% each of which is referred to as $D_p$, where $p = 1,..., N$. %\textbf{Figure} displays exemplary examples of $D_p$ from both the real retail product image dataset $D_r$ and the simulated retail product image dataset $D_s$. 
Note that different products in the dataset have different colors, orientations, physical and pixel dimensions. We don't assume the presence of a classification label for each product. The goal of this module is to extract features corresponding to each product, that can be used to perform context-based anomaly detection.

In the retail domain, collecting labels for every product may not be feasible in all cases and because of constant change in product packaging, there is a high probability that a model might not be able to classify different products into their respective categories. Additionally, it has been observed that even within the same class, there exist variations in visual appearance based on different flavors (colors of packaging), sizes, etc. Therefore, it might not be the best strategy to train a classifier using class labels and then use this classifier to extract discriminative features corresponding to each product in order to perform anomaly detection. As a result of these factors, an unsupervised technique is required to get meaningful features that can be utilised to establish contexts for context-based anomaly identification. 
For instance, an anomaly can be color-based when a red colored packet is placed on a shelf among blue packets or it can be shape-based when a ball is placed among Rubik's cubes.

One possible solution which is extensively explored for the planogram compliance problem is to use handcrafted features like key-point-based (Scale-Invariant Feature Transform (SIFT)~\cite{mikolajczyk2004scale}, Speeded-Up Robust Feature (SURF)~\cite{bay2006surf}), gradient-based (Histogram of Oriented Gradients (HOG), Sobel, Prewitt operators), color-based (color histogram), pattern-based (Haar-like features). However, these techniques do not always display sufficient information and it becomes an engineering effort to choose the features and parameters that fit best. To deal with this, we can use deep learning (DL)-based solutions which have shown improved performance. Thus, we propose an auto-encoder architecture which can learn disentangled color and shape content latent embeddings as in~\cite{sengar2022low}. 

We compare the performance of our method with other baselines and show that the disentangled latent embeddings are useful for identifying the anomalies in an explainable manner. Further, the advantage in having disentangled latent embeddings over distributed embeddings is that only important and necessary subsets of the embeddings can be utilized for a downstream task. For example, a robotic gripper requires information about the shape, size and texture of the surface of the object to infer a grasp but does not require color information. As a result, by providing relevant information to the gripper, one can increase the signal-to-noise ratio in the input space. This is not possible to do in conventional auto-encoder setups since
they contain all the necessary information to accurately reconstruct the original image, but the information related to different concepts like shape, size, location, color, etc. are all in a tightly entangled representation. This entangled information is not discriminative enough to perform context based anomaly detection. Also, the planning module typically acts on local representations of the concepts but DL models give distributed representations where concepts are denoted by continuous-value vectors~\cite{garcez2020neurosymbolic}. Thus, in order to bridge between the two paradigms the first step is to disentangle the representations given by standard DL models without the use of external concept annotations. 

In order to capture disentangled color and content information, we introduce two methods wherein we train a vision transformer (ViT)~\cite{vaswani_attention_2017, dosovitskiy_image_2020} to encode images into latent embeddings. The two methods differ in how the shape and texture information (content information) of the object is captured in the latent representation. In the first method, ViT concept mining DWT (ViT-CM-DWT), we use Discrete Wavelet Transform (DWT) using Haar as the mother wavelet, with content latent embeddings specialized to reconstruct the DWT components of the gray scale image of the object (e.g., Level-1 DWT components: LL, HL, LH, and HH) similar to \cite{sengar2022low}. In the second method, ViT concept mining (ViT-CM), we extract content latent embeddings to reconstruct the gray scale image of the object without using DWT. On the other hand, the color latent embeddings are trained to capture the color information of an image.

The ViT auto-encoder (ViT-CM-DWT) architecture for color and shape content feature specialisation is shown in Fig.~\ref{fig:architecture}. The RGB images used to train the ViT-AE are of size 3x224x224 (CxHxW). The Linear Projection layer is made up of a single convolutional network (input channels=3, output channels=128, kernel size = 16 and patch size = 16) which takes in an object image and implicitly splits them into image patches of size 3x16x16 to generate a feature vector of 128 dimension for each image patch. Further, the latent embeddings are split into two equal halves of dimension 64 each. One part of the embedding is passed through the 4 shape transformer encoder layers (one for each DWT component LL, HL, LH, HH) and the other half is passed to the 3 color transformer encoder layers (one for each Red channel, Green channel and Blue channel). Each transformer encoder comprises of 4 attention heads, a feed forward network with a hidden dimension of 2048 and uses GELU activation function. The intention is to specialize a part of the embedding to learn the different shapes and textures present in the object image and the other part to learn the colors. The position embeddings are only made available to the latent embeddings for content. The latent embedding $f\in R^{N\times N_C\times M}$ of an input image $I_{c}\in R^{H\times W\times 3}$ is produced by the encoder network ($E$). 
\begin{equation}
\label{eq:feature_extraction}
f = E(I_{c};\theta_{E}) 
\end{equation}
Here, latent embedding $f\in R^{N\times N_C\times M}$ comprises of disentangled color and content specialized features where $N$ represents the total number of patches ($N$=16), $N_{C}$ is the total number of specialized features or concepts ($N_{C}$=7 because there are 3 color channels and 4 DWT components) and $M$ denotes the dimension of these features.
\begin{align*}
&f = \{f_{color},f_{content}\} \\
&f_{color} = \{f_{R},f_{G},f_{B}\}\\
&f_{content} = \{f_{LL},f_{HL},f_{LH},f_{HH}\} 
\end{align*}
Similar to \cite{sengar2022low}, the color embeddings $f_{R}$, $f_{G}$ and $f_{B}$ are used to modulate the content embeddings $f_{content}$, as described below:

\begin{eqnarray}
\label{eq:mod_eq}
f_{content}^{R} = \{f_{LL}\odot f_{R},f_{HL} \odot f_{R},f_{LH} \odot f_{R},f_{HH}\odot f_{R}\} \\
f_{content}^{G} = \{f_{LL}\odot f_{G},f_{HL} \odot f_{G},f_{LH} \odot f_{G},f_{HH}\odot f_{G}\} \\
f_{content}^{B} = \{f_{LL}\odot f_{B},f_{HL} \odot f_{B},f_{LH} \odot f_{B},f_{HH}\odot f_{B}\} 
\end{eqnarray}

Here, $\odot$ represents the Hadamard product and $f_{LL}$, $f_{HL}$, $f_{LH}$ and $f_{HH}$ are the specialized features for each of the DWT components. Following this, the modulated features are passed to the decoder bank to reconstruct the DWT components of the input image $I_{c}$.
\begin{eqnarray}
\phi_{DWT}^{x}  &=& D(f_{content}^{x};\theta_D)  \end{eqnarray}
Where, $x\in\{R,G,B\}$, $\theta_D$ represent the parameters of decoder $D$. The image is reconstructed back using the IDWT module~\cite{cotter2020uses} through the standard process.

The auto-encoder is trained in an end-to-end fashion with squared L2 loss between the input image and the reconstructed image. Note that to disentangle the color and content information, we train the network alternatively where in one iteration the $f_{content}$ embeddings are modulated with $f_{color}$ embeddings as described in equations (2)-(4) above. $f_{content}$ is detached from the computational graph so that it does not learn to contain information about the color of the object image.
In this case, the L2 loss is calculated between the RGB image and the reconstructed image. In the next iteration, $f_{content}$ embeddings are passed as is, without modulation, enforcing the network to only learn the content features and no color information is passed to the decoder for reconstruction. In this case, the loss is computed between gray-scale input image and the reconstructed image. %To check whether the embeddings are disentangled we have shown color and content interpolation results in \textbf{Figure}.

In case of ViT-CM, we replace the 4 content encoder layers with a single content encoder layer to learn the shape and texture information of the object. The training process remains the same except, we have a single content latent embedding $f_{content}$ instead of 4.

To train our models and baselines we use the object RGB images and segmentation masks in the YCB dataset \cite{ycb-dataset}. We train our models for 100 epochs with a learning rate of 1e-4 using Adam optimizer and mean squared error loss.

\subsection{Algorithm for Anomaly Detection}

The features obtained from the concept embedding extraction modules are then used to perform anomaly detection using outlier detection algorithms. We implemented two such algorithms. In the first one, we applied agglomerative clustering on the concept embeddings corresponding to each cropped image. In the second one, we calculated pairwise distances between the embeddings corresponding to each cropped image and then computed a row-wise sum on the resulting distance matrix. On the aggregated distance we then detected outliers based on the inter-quartile range. 

\begin{algorithm}
	\caption{Anomaly Detection} 
	\begin{algorithmic}[1]
		\For {$evaluation\_set=1,2,\ldots,K$}
			\For {$object\_images=1,2,\ldots,N$}
				\State Compute feature vectors $\hat{f}_{1},\ldots,\hat{f}_{N}$ using Eq~\ref{eq:feature_extraction}
			\EndFor
			\State Compute distance between feature vectors using a similarity metric
			\State $anomaly[K]=argmax(distance)$
		\EndFor
	\end{algorithmic} 
\end{algorithm}

% \textbf{To add pseudo code for overall algorithm}. 

\section{Evaluation}
\label{sec:results}
% Add more details, add unsuccessful examples

We have evaluated our proposed approach for anomaly detection on a simulated dataset as well as on publicly available retail product dataset, RP2K. To evaluate our anomaly detection approach, we have created evaluation sets by randomly choosing majority samples from one class and anomalous sample from another randomly chosen class. 
The Co-AD approaches tested here include ViT-CM and ViT-CM-DWT architectures with selections of content features, color features, or both.
These are compared with a similar architecture that uses convolutional neural networks (CNN) for concept mining to generate latent embeddings (CNN-CM-DWT) with the same feature selection choices~\cite{sengar2022low}.
We have also compared our Co-AD approach with standard ViT auto-encoders which do not need product labels, as well as with a deep residual network (ResNet50)~\cite{he2016deep} that requires labelled data.
Co-AD was found to generate anomaly detection outputs from a given simulated or real scene at $\sim$5 fps on the onboard computer of the mobile robot (NVIDIA Jetson TX2).

\subsection{Image Dataset - RP2K}
\label{subsec:rp2k_dataset}
The RP2K dataset~\cite{rp2k} contains two components: the original shelf images and the individual object images cropped from the shelf images. All images are captured in physical retail stores with natural lightings, matching the scenario of real applications. In this dataset each individual object is at least 80 by 80 pixels. To prepare the evaluation set, we considered cropped images from the dataset. %Representative images from the dataset are shown in Fig~\ref{fig:rp2k_set}. 
There are a total of 198 test cases in the evaluation set. The anomaly detection accuracy is listed in Table~\ref{tab:RP2k_res}.
ViT-CM-DWT with color features and a pairwise boxplot distance metric had the highest accuracy for this dataset (89.9\%).

\begin{table}[]
\begin{tabular}{|c|c|c|}
\hline
\multicolumn{1}{|c|}{Model} &
  \begin{tabular}[c]{@{}c@{}}Agglomerative\\ clustering on\\ features\end{tabular} &
  \begin{tabular}[c]{@{}c@{}}Boxplot on\\ pairwise\\ distance\end{tabular} \\ \hline
\begin{tabular}[c]{@{}c@{}}ResNet\end{tabular} & 51.52\% & 69.7\% \\ \hline
\begin{tabular}[c]{@{}c@{}}ViT-AE\end{tabular} & 74.75\% & 80.81\% \\ \hline
%\begin{tabular}[c]{@{}c@{}}CNN-AE\end{tabular} & 6.06\% & 0.51\% \\ \hline
\begin{tabular}[c]{@{}c@{}}CNN-CM-DWT\\ (Color features)\end{tabular} & 29.8\% & 43.43\% \\ \hline
\begin{tabular}[c]{@{}c@{}}CNN-CM-DWT\\ (Content features)\end{tabular} & 46.97\% & 56.06\% \\ \hline
\begin{tabular}[c]{@{}c@{}}CNN-CM-DWT\\ (Content \& Color features)\end{tabular} & 48.48\% & 59.6\% \\ \hline
\begin{tabular}[c]{@{}c@{}}ViT-CM-DWT\\ (Color features)\end{tabular} & 84.34\% & 89.9\% \\ \hline
\begin{tabular}[c]{@{}c@{}}ViT-CM-DWT\\ (Content features)\end{tabular} & 41.41\% & 45.45\% \\ \hline
\begin{tabular}[c]{@{}c@{}}ViT-CM-DWT\\ (Content \& Color features)\end{tabular} & 83.84\% & 86.87\% \\ \hline
\begin{tabular}[c]{@{}c@{}}ViT-CM\\ (Color features)\end{tabular} & 76.77\% & 84.85\% \\ \hline
\begin{tabular}[c]{@{}c@{}}ViT-CM\\ (Content features)\end{tabular} & 57.58\% & 65.66\% \\ \hline
\begin{tabular}[c]{@{}c@{}}ViT-CM\\ (Content \& Color features)\end{tabular} & 80.3\% & 89.39\% \\ \hline
\end{tabular}
\caption{Success rate on RP2K dataset}
\label{tab:RP2k_res}
\end{table}

\subsection{Simulated Images}
\label{subsec:simulated_dataset}
We also evaluate Co-AD on a set of images containing objects arranged on shelves in a ROS Gazebo simulation environment. As shown in Fig.~\ref{fig:shelf_localization}, the test set contains images of 12 different object classes drawn from the YCB dataset~\cite{ycb-dataset}. We have created 72 test cases containing different anomalous objects. Table~\ref{tab:sim_results} lists the accuracy on simulated data.
ViT-CM with color features and a pairwise boxplot distance metric had the highest accuracy for this dataset (95.83\%).

\begin{table}[]
\begin{tabular}{|c|c|c|}
\hline
\multicolumn{1}{|c|}{Model} &
  \begin{tabular}[c]{@{}c@{}}Agglomerative\\ clustering on\\ features\end{tabular} &
  \begin{tabular}[c]{@{}c@{}}Boxplot on\\ pairwise\\ distance\end{tabular} \\ \hline
\begin{tabular}[c]{@{}c@{}}ResNet\end{tabular} & 75.0\% & 88.89\% \\ \hline
\begin{tabular}[c]{@{}c@{}}ViT-AE\end{tabular} & 72.22\% & 70.83\% \\ \hline
%\begin{tabular}[c]{@{}c@{}}CNN-AE\end{tabular} & 0.0\% & 0.0\% \\ \hline
\begin{tabular}[c]{@{}c@{}}CNN-CM-DWT\\ (Color features)\end{tabular} & 50.0\% & 62.5\% \\ \hline
\begin{tabular}[c]{@{}c@{}}CNN-CM-DWT\\ (Content features)\end{tabular} & 47.22\% & 50.0\% \\ \hline
\begin{tabular}[c]{@{}c@{}}CNN-CM-DWT\\ (Content \& Color features)\end{tabular} & 55.56\% & 69.44\% \\ \hline
\begin{tabular}[c]{@{}c@{}}ViT-CM-DWT\\ (Color features)\end{tabular} & 86.11\% & 90.28\% \\ \hline
\begin{tabular}[c]{@{}c@{}}ViT-CM-DWT\\ (Content features)\end{tabular} & 40.28\% & 43.06\% \\ \hline
\begin{tabular}[c]{@{}c@{}}ViT-CM-DWT\\ (Content \& Color features)\end{tabular} & 84.72\% & 88.89\% \\ \hline
\begin{tabular}[c]{@{}c@{}}ViT-CM\\ (Color features)\end{tabular} & 80.56\% & 95.83\% \\ \hline
\begin{tabular}[c]{@{}c@{}}ViT-CM\\ (Content features)\end{tabular} & 45.83\% & 47.22\% \\ \hline
\begin{tabular}[c]{@{}c@{}}ViT-CM\\ (Content \& Color features)\end{tabular} & 93.06\% & 91.67\% \\ \hline
\end{tabular}
\caption{Success rate on Simulated dataset}
\label{tab:sim_results}
\end{table}

\begin{figure}[ht]
\centering
\includegraphics[width=0.65\columnwidth]{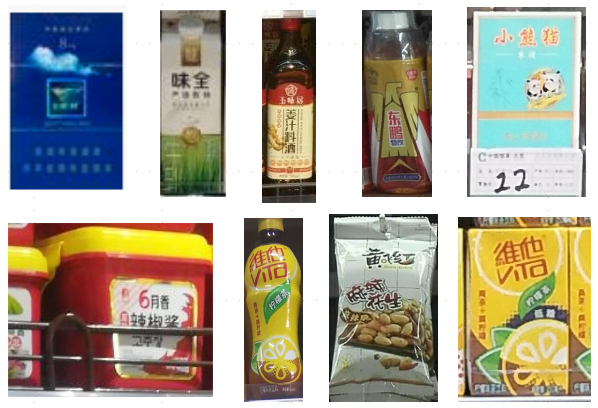}
\caption{Representative images of retail objects present in the RP2K dataset~\cite{rp2k}}
\label{fig:rp2k_set}
\end{figure}

\subsection{Evaluation Sets - Failure Cases}
\begin{figure}[ht]
\centering
\includegraphics[width=0.65\columnwidth]{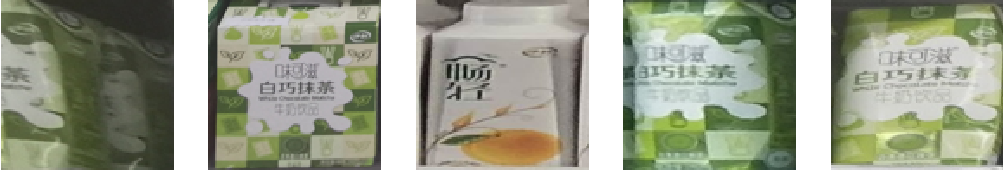}
\includegraphics[width=0.65\columnwidth]{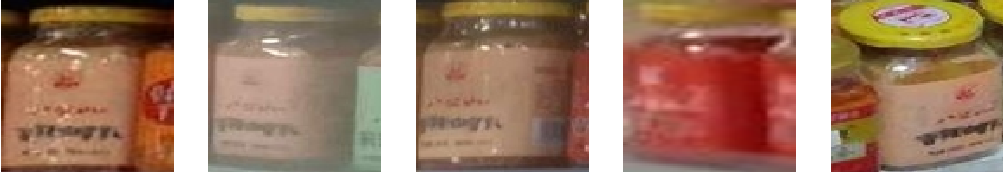}
\caption{Examples of images from the RP2K~\cite{rp2k} evaluation set where anomaly detection using ViT-CM fails due to color features (top row) and content features (bottom row).}
\label{fig:rp2k_eval_set_fail}
\end{figure}

\begin{figure}[ht]
\centering
\includegraphics[width=0.7\columnwidth]{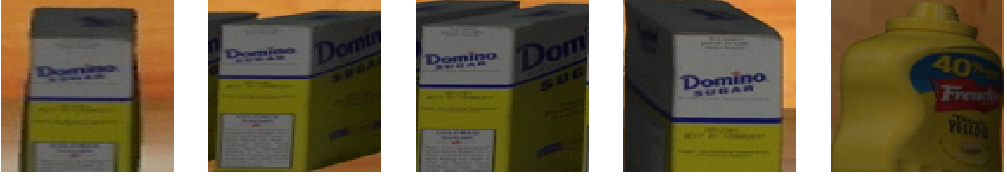}
\includegraphics[width=0.7\columnwidth]{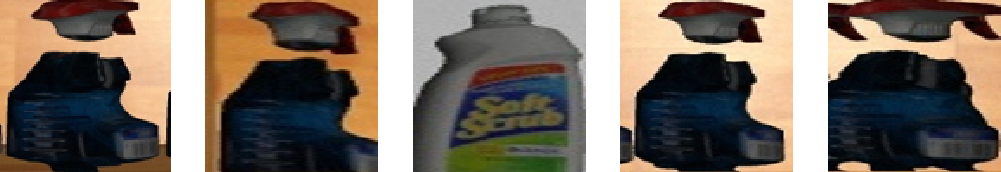}
\caption{Examples from the simulated images~\cite{ycb-dataset} evaluation set where anomaly detection using ViT-CM fails due to color features (top row) and content features (bottom row).}
\label{fig:simulated_eval_set_fail}
\end{figure}

To demonstrate the limitations of Co-AD approaches, we present two examples of evaluation sets from the simulated images and RP2K images where ViT-CM fails to identify an anomaly.
The failure cases are identified after using agglomerative clustering on both color and content features. In the first row of both Fig.~\ref{fig:rp2k_eval_set_fail} and Fig.~\ref{fig:simulated_eval_set_fail}, the ViT-CM approach fails because the color feature vectors of each of the object cannot be told apart. On the other hand, the second row of both Fig.~\ref{fig:rp2k_eval_set_fail} and Fig.~\ref{fig:simulated_eval_set_fail} we observe that the content features of all the object images are close to each other that leads to inaccurate anomaly detection. 

The Concept-based Anomaly Detection (Co-AD) approach presented in this paper consists of various design choices-- inclusion of DWT, choice of color and content features and choice of outlier detection algorithm. As seen in the evaluation results, the anomaly detection performance on different test sets varies with these choices. Co-AD can therefore be suitably modified and adapted to a particular application area through these design choices with minimal changes in the underlying architecture.

\section{Automatic Anomaly Correction Demos}
\label{sec:demo}

While the concept-based anomaly detection methods outperformed baselines on the evaluation image datasets, their real utility is as part of an autonomous robotics pipeline, obviating the need for human intervention.%, as described in  Sec.~\ref{sec:system_desc}-C.

%The overall concept for achieving this goal is described in Sec.~\ref{sec:system_desc}-C, with the anomaly detector playing a key role in the robot's planing pipeline. In this section, we present two demonstrations of the anomaly detection and correction pipeline--- in a ROS Gazebo simulation environment, and on a physical mobile manipulator robot. 

We demonstrate the pipeline on a mobile manipulation platform (Fig.~\ref{fig:intro_fig}) consisting of a custom mobile base~\cite{mopicker_patent}, a Universal Robots UR5e arm, an OnRobot RG6 gripper (Robotiq 2F-85 in simulation), and an Intel Realsense D415 depth camera.

\subsection{ROS Gazebo Simulator}

\begin{figure}[ht]

\begin{subfigure}[b]{0.49\columnwidth}
         \centering
         \includegraphics[width=0.75\textwidth]{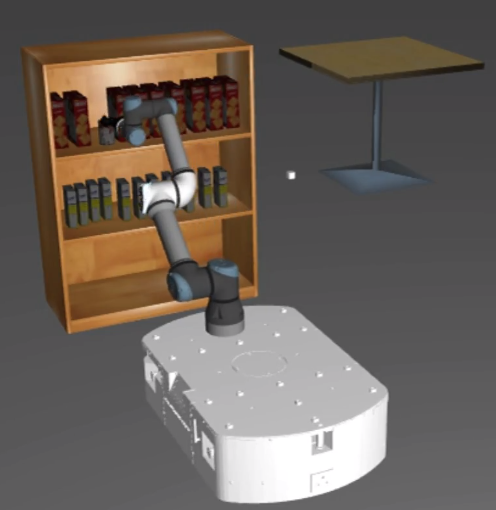}
         \caption{Simulation environment}
         \label{fig:sub:sim_env}
\end{subfigure}
\begin{subfigure}[b]{0.49\columnwidth}
         \centering
         \includegraphics[width=0.75\textwidth]{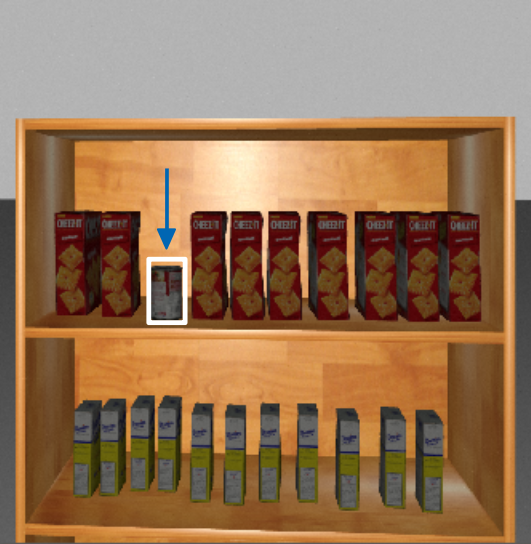}
         \caption{RGB image}
         \label{fig:sub:sim_cam}
\end{subfigure}

\caption{An illustrative task simulated in a ROS Gazebo environment: (a) mock retail setup, (b) image captured by the robot's on-body camera with anomalous object shown.}
\label{fig:gazebo}
\end{figure}

The ROS Gazebo simulation of a retail anomaly detection task consists of a shelf with objects placed on two rows. The objects are drawn from the YCB dataset~\cite{ycb-dataset} of everyday objects.
The depth camera is simulated via a Gazebo plugin and placed on the robot's body at roughly the same position as the real camera.
The robot's objective is to scan the shelf, flag the anomalous object using Co-AD, pick it up, and place it on the table behind the shelf that serves as an inventory buffer area (Fig.~\ref{fig:gazebo}a) whose location is known \textit{a priori} in the world frame.

Following the procedure described in Sec.~\ref{sec:anom_det}, the soup can is identified as the anomaly (Fig.~\ref{fig:gazebo}b).
This triggers the pipeline shown in Fig.~\ref{fig:schematic}, calling a PDDL planner that generates a task plan with the primitive actions \textit{move}, \textit{pick-up} and \textit{put-down}, for taking the can from the shelf to the table.
The task plan is parsed into sequence of motion plans using Moveit~\cite{moveit} for planning the robotic arm's trajectory and ROS Navigation for the mobile base.

\subsection{Physical Demonstration}

\begin{figure}[ht]
\begin{subfigure}[b]{0.49\columnwidth}
         \centering
         \includegraphics[width=0.75\textwidth]{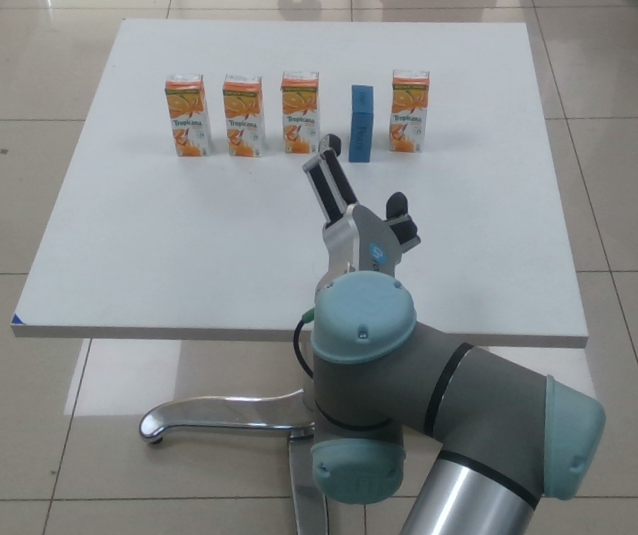}
         \caption{RGB image}
         \label{fig:sub:real_rgb}
\end{subfigure}
\begin{subfigure}[b]{0.49\columnwidth}
         \centering
         \includegraphics[width=0.65\textwidth]{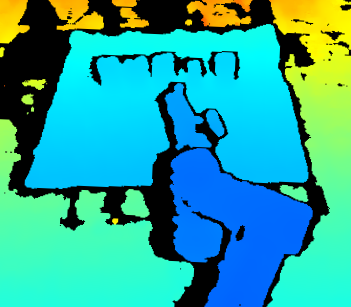}
         \caption{Depth map}
         \label{fig:sub:real_depth}
\end{subfigure}

\begin{subfigure}[b]{0.49\columnwidth}
         \centering
         \includegraphics[width=0.7\textwidth]{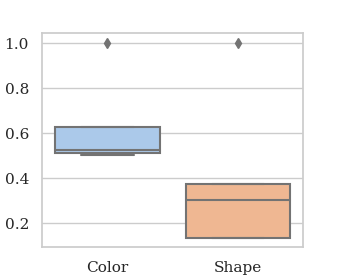}
         \caption{Boxplots in feature space}
         \label{fig:sub:real_cluster}
\end{subfigure}
\begin{subfigure}[b]{0.49\columnwidth}
         \centering
         \includegraphics[width=0.7\textwidth]{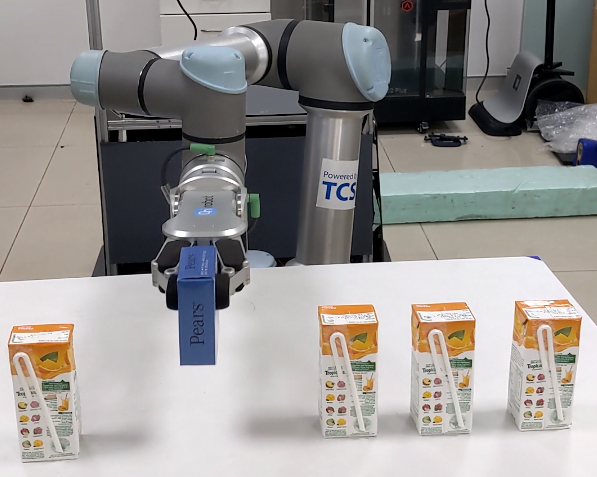}
         \caption{Picking target object}
         \label{fig:sub:real_pick}
\end{subfigure}

\caption{A simplified task executed on the robot with anomaly detection: (a) image from robot's on-body camera, (b) corresponding depth image, (c) boxplots of normalized pairwise distance in shape content and color features with the outlier shown, (d) anomalous object picked up.}
\label{fig:real_imgs}
\end{figure}

Another demonstration was conducted with the physical robot, with one object of a different class placed among four other objects (Fig.~\ref{fig:real_imgs}).
The objects were picked from a local grocery store.
The anomalous object was detected using Co-AD, and picked up with the robotic arm.
Fig.~\ref{fig:sub:real_cluster} shows boxplots for Co-AD concept embeddings for the five objects in the image. 
The similarity metric is normalized pairwise distances in the color features and shape content features.
The dissimilar object can be seen as an outlier.

\section{Conclusion}
\label{sec:conclusion}
In this paper, we presented a concept-based anomaly detection method (Co-AD) that allows for the detection of misplaced items in a retail store without relying on planograms or labeled object databases. While this approach performed well on real and simulated image datasets, investigation and improvement of its real-world performance in physical retail stores remains ongoing.

As part of a mobile manipulation platform that can autonomously correct anomalies in retail stores, Co-AD is useful due to its scalability and low computational burden. While the current implementation uses well-established techniques for task and motion planning, a real deployment requires more reactivity and adaptability to changing environments in terms of navigation and manipulation. These challenges have been enumerated in~\cite{sengar2022challenges} and the development of solution strategies constitutes ongoing and future work.

\bibliographystyle{IEEEtran}
\bibliography{IEEEabrv,refs}

\end{document}